\documentclass{article}
\pdfoutput=1
\usepackage{times}
\usepackage{graphicx} 

\usepackage[square,sort,comma,numbers]{natbib}

\usepackage{algorithm}
\usepackage{algorithmic}
\usepackage{amsthm}

\usepackage{hyperref}

\usepackage{url}
\usepackage{caption}
\usepackage{subcaption}
\usepackage{amsmath}

\usepackage[final]{nips_2018}

\usepackage[utf8]{inputenc} 
\usepackage[T1]{fontenc}    
\usepackage{hyperref}       
\usepackage{url}            
\usepackage{booktabs}       
\usepackage{amsfonts}       
\usepackage{nicefrac}       
\usepackage{microtype}      

\newcommand{\esm}[1]{\ensuremath{#1}}

\newcommand{\ms}[1]{\esm{\mathsf{#1}}}

\newcommand\reals{\ms{R}}

\newcommand\sparam{\alpha}

\newcommand\synteq{::=}

\newcommand\integratedgrads{\ms{IG}}
\newcommand\internalinfluence{\ms{IntInf}}
\newcommand\conductance{\ms{Cond}}

\newcommand\xbase{x'}

\title{How Important Is a Neuron?}

\author{
  Kedar Dhamdhere \\
  Google AI\\
  \texttt{kedar@google.com} \\
  \And
  Mukund Sundararajan \\
  Google AI\\
  \texttt{mukunds@google.com} \\
   \AND
  Qiqi Yan \\
  Google AI\\
  \texttt{qiqiyan@google.com} \\
}
\begin{document}

\maketitle



\begin{abstract}
  The problem of attributing a deep network's prediction to its \emph{input/base} features is well-studied (cf.~\cite{SVZ13}).
  We introduce the notion of \emph{conductance} to extend the notion of attribution to the understanding the importance of \emph{hidden} units.

  Informally, the conductance of a hidden unit of a deep network is the \emph{flow} of attribution via this hidden unit.
  We use conductance to understand the importance of a hidden unit to the prediction for a specific input, or over a set of inputs. We evaluate the effectiveness of conductance in multiple ways, including theoretical properties, ablation studies, and a feature selection task. The empirical evaluations are done using the Inception network over ImageNet data, and a sentiment analysis network over reviews. In both cases, we demonstrate the effectiveness of conductance in identifying interesting insights about the internal workings of these networks.
\end{abstract}

\section{Background and motivation}
\label{sec:motivation}

The problem of attributing a deep network's prediction to its input is well-studied (cf.~\cite{BSHKHM10,SVZ13,SGSK16,BMBMS16,SDBR14,LL17, STY17}).
Given a function $F: \reals^n \rightarrow [0,1]$
  that represents a deep network, and an input $x = (x_1,\ldots,x_n) \in \reals^n$.
  An attribution of the prediction at input $x$ is $(a_1,\ldots,a_n) \in \reals^n$
  where $a_i$ can be interpreted as the  \emph{contribution} of $x_i$ to the
  prediction $F(x)$. For instance, in an object
recognition network, an attribution method could tell us which pixels
of the image were responsible for a prediction. Attributions help determine the influence of base features on the input.

Many papers on deep learning speculate about the importance of a hidden unit towards a prediction. They tend to use the activation value of the hidden unit or its product with the gradient as a proxy for feature importance. As we discuss later in the paper, both measures have undesirable behavior. For instance, a ReLU always has positive activation values but could either have positive or negative influence on the prediction; the fact that the sign of the influence cannot be identified is undesirable. A recent paper~\cite{influence} is closely related to ours in intention and technique. It proposes a notion of influence of an hidden neuron on the output prediction. We explicitly compare our technique to theirs, both qualitatively and quantitatively. We show for instance, that their technique can sometimes result in importance scores with incorrect signs.

  There is also a large literature on understanding the function/operation of hidden units of a deep network (cf.~\cite{MV15, DB15, YCNFL15, deepdream, featurevis}). These approaches are usually optimization based, i.e. they explicitly tweak images to optimize activations of a neuron or a group of neurons.  The resulting images are an indicator of the function of the hidden unit. However, these works don't identity the importance of a hidden unit (either to a prediction or to a class of predictions). Also, these approaches tend not to work for every hidden unit. Consequently, it is not clear if all the hidden units which are important to the predictive task have been investigated by these approaches. Another approach to identifying the function of hidden units is to build a simple, linear, explanatory model (cf.~\cite{QVL13,probe}) in parallel to the original model. But then it is no longer clear if the influence of the hidden unit in the new model is the same as what it is in the original model; the nature of the correlations in the two models may differ. Finally, ~\cite{kim} takes a human-defined concept as input, finds hidden neurons that are predictive of this concept, and identifies the influence of these neurons on a certain prediction. Because the network may rely on an inherently different representation from the human, it is not clear how this process identifies all the concepts important to the prediction for a given input or a class of inputs. In contrast, we rely on the network's own abstraction, i.e., the filters in the hidden layers. This will result in accounting for the whole prediction, but may possibly result in less intuitive explanations.

\section{Our Contribution}
\label{sec:contribution}

We introduce the notion of conductance (see  Section~\ref{sec:conductance}) based on a previously studied attribution technique called Integrated Gradients~\cite{STY17}. Integrated Gradients rely on integrating (summing) the gradients (of the output prediction with respect to the input) over a series of carefully chosen variants of the input. Informally, the conductance of a hidden unit of a deep network is the \emph{flow} of Integrated Gradients' attribution via this hidden unit. The key idea behind conductance is to decompose the computation of Integrated Gradients via the \emph{chain rule} (see Equations~\ref{eq:pixel-conductance} and~\ref{eq:total-conductance} for formal definitions).

We evaluate the effectiveness of conductance (against several related methods) in several ways. We identify the nice theoretical properties that conductance has. We use ablations to see if the removal of high-conductance units would change predictions. We do a feature selection study to see if the high-conductance units have strong predictive power using transfer learning. We also evaluate by seeing if conductance can give us intuitive insights about the network. Our empirical evaluations are done over the following two large-scale networks.

For the widely used Inception architecture~\cite{SLJSRAEVR14} trained on ImageNet data~\cite{ILSVRC15}, for every input, we are able to find a small number of high-conductance filters ($5$ to $10$ out of $5484$ filters) that account for the classification for the input. We verify that removing these high-conductance filters would alter classification output. Moreover, some of these filters are influential \emph{across} images and labels; these seem to correspond to texture or color features. We also find that filters that have high conductance on average across images from a single class, are highly predictive of the class.

The second one is a highly-cited convolution network~\cite{K14} frequently used for text classification tasks, trained over review data to perform sentiment analysis. Unlike the object recognition network, we find that a relatively large number of filters are influential for every input, i.e., the behavior of the network is redundant. Furthermore, we find that there is a division of labour between the filters.  Most filters contribute to either positive sentiment, or negative sentiment, but not both. Finally, we identify the hidden units capturing negation.


\section{Conductance}
\label{sec:conductance}

We start with some notation. Formally, suppose we have a
function $F: \reals^n \rightarrow [0,1]$ that represents a deep network.
Specifically, let $x \in \reals^n$ be the input at hand, and
$\xbase \in \reals^n$ be the baseline input. For image networks, the baseline
could be the black image, while for text models it could be the zero
embedding vector. \footnote{Several attribution methods(cf.~\cite{SGSK16, BMBMS16, STY17}) produce attributions \emph{relative} to an extra input called the baseline.
  All attribution methods rely on some form of sensitivity analysis, i.e., perturbing the input or the network, and examining how the prediction changes with such perturbations.
  The baseline helps defines these perturbations.}


We now recall the definition of Integrated Gradients. We consider the straightline path (in $\reals^n$) from the baseline $\xbase$ to the input
$x$, and compute the gradients at all points along the path. Integrated gradients
are obtained by cumulating these gradients. Specifically, integrated gradients
are defined as the path intergral of the gradients along the straightline
path from the baseline $\xbase$ to the input $x$.
Here is the intuition for why the method works for, say, an object recognition network.
The function $F$ varies from a near zero value for the informationless baseline to its final value.
The gradients of $F$ with respect to the image pixels explain each step of the variation in the value of $F$.
The integration (sum) over the gradients cumulates these micro explanations and accounts for the net difference between the baseline prediction score (near zero)
and the prediction value at the input $x$.

Formally, the integrated gradient for the  $i^{th}$ base feature (e.g. a pixel) of an input $x$ and baseline
$\xbase$ is:
\begin{equation}
\label{eq:int-grads}
\integratedgrads_i(x) \synteq (x_i-\xbase_i)\cdot\int_{\sparam=0}^{1} \tfrac{\partial F(\xbase + \sparam (x-\xbase))}{\partial x_i  }~d\sparam
\end{equation}
where
$\tfrac{\partial F(x)}{\partial x_i}$ is the gradient of
$F$ along the $i^{th}$ dimension at $x$.

Notice that Integrated Gradients produces attributions for base features (e.g. the pixels of an object recognition network). 
There is a natural way to `lift' these attributions to a neuron in a hidden layer.
Consider a specific neuron $y$ in a hidden layer of a network. We can define the conductance of neuron $y$ for the attribution to an input variable $i$ as:
\begin{equation}
  \label{eq:pixel-conductance}
\conductance_i^{y}(x) \synteq (x_i-\xbase_i)\cdot\int_{\sparam=0}^{1} \tfrac{\partial F(\xbase + \sparam (x-\xbase))}{\partial y} \cdot \tfrac{\partial y}{\partial x_i} ~d\sparam 
\end{equation}

We can define the total conductance of the hidden neuron $y$ by summing over the input variables:
\begin{equation}
\label{eq:total-conductance}
\conductance^{y}(x) \synteq \sum_{i}  (x_i-\xbase_i)\cdot\int_{\sparam=0}^{1} \tfrac{\partial F(\xbase + \sparam (x-\xbase))}{\partial y} \cdot \tfrac{\partial y}{\partial x_i} ~d\sparam 
\end{equation}

We will interchangably use the term ``conductance of neuron $j$'' to mean either of Equations~\ref{eq:pixel-conductance} or~\ref{eq:total-conductance}.
We use the former to explain the function of the neuron in terms of its effect on the base features of the input.
We use the latter to discuss the importance of the neuron.

Frequently, we will aggregate over a set of logically related neurons that belong to a specific hidden layer.
For instance, these could be neurons that belong to a single filter. 
In this case, we will sum over the conductances of the neurons in the set to define the conductance of the set as whole.

\section{Evaluation of Conductance}

We compare conductance against three other methods. The first two are commonly used in literature and the third is from a recent paper~\cite{influence}.

\begin{itemize}
\item \textbf{Activation:} The value of the hidden unit is the feature importance score.
\item \textbf{Gradient*Activation:} $y \times \tfrac{\partial F(\xbase + \sparam\times(x-\xbase))}{\partial y}$
\item \textbf{Internal Influence:}  The measure of feature importance is:
  \begin{equation}
  \internalinfluence^{y}(x) \synteq \int_{\sparam=0}^{1} \tfrac{\partial F(\xbase + \sparam\times(x-\xbase))}{\partial y}  ~d\sparam
  \end{equation}
\end{itemize}

It is notoriously hard to distinguish strange model behavior from a bad feature importance technique; this affects all of the literature on attribution.
We provide three types of evaluation to get around this. We compare theoretical properties of the different methods (this section). Second, we contrast the importance scores against the change in score via ablations. The premise of this evaluation is that removing a hidden unit
should have a somewhat commensurate impact on the ablation score (cf.~\cite{SingleDirection}) (This is going to be somewhat distorted by the non-linearity of the network, because the ablations are done one at a time. As we discuss in Section~\ref{sec:object-eval}, ablations artifactually favor gradient*activation.).As we show in Section~\ref{sec:object} and Section~\ref{sec:sentiment}, conductance have a high degree of agreement with ablation scores. Third, we use the importance scores within a feature selection task. The premise is that hidden units that are important across a set of inputs from a class should be predictive of this input class. Again, we find that conductance is superior to the other methods.

\subsection{Properties of Conductance}
\label{sec:properties}

  Conductance is a natural refinement of Integrated Gradients.
  Consequently, several properties of Integrated Gradients carry over to conductance.
  Here, we mention some relevant properties.

  Conductances satisfy a property called \emph{completeness} which means that the conductances for any single hidden layer add up to the difference between the predictions $F(x) - F(\xbase)$. \footnote{The proof follows from the fact that Integrated Gradients satisfies completeness and   the fact that derivative $\frac{\partial F}{\partial x_i}$ can be expressed using the chain rule as the sum $\sum_{j} \frac{\partial F}{\partial y_j} \cdot \frac{\partial y_j}{\partial x_i}$, where $j$ indexes the neurons in the layer.} An immediate consequence is that conductances thus satisfy the \emph{Layerwise Conservation Principle}~\cite{BBMKMS15}, which says that ``a network's output
  activity is fully redistributed through the layers of a DNN onto the input variables, i.e., neither positive nor negative evidence is lost.''~\cite{SBMBM15}. None of the three methods we compare against satisfy completeness or layerwise conservation (the bad examples from the previous section also show this).

  Conductance satisfies \emph{linearity}. (So do internal influence and gradient*activations.) Suppose that we linearly compose hidden neurons $f_1$ and $f_2$ to form the final network that models the
  function $a\times f_1 + b\times f_2$. Then, the conductances of the two hidden neurons will be $a\times (f_1(x) - f_1(\xbase))$ and $b\times (f_2(x) - f_2(\xbase))$ respectively. This is a sanity-check because if the action of a network is mostly linear from a hidden layer, the conductances will match what is intuitively the obvious solution.

  Conductance is also appropriately \emph{insensitive}. If varying the values of a hidden unit does not change the network's prediction, it has zero conductance. If varying the inputs does not change value of the hidden unit, the hidden unit has zero conductance. (The proofs for both properties are based on the derivatives $\frac{\partial F}{\partial y_j}$ or $\frac{\partial y_j}{\partial x_i}$ being zero.)

  Though we do not provide an axiomatic result like~\cite{STY17} that the way we define conductance is the unique best way to do so, we assert that our definition of conductance is mathematically natural (decomposing Integrated Gradients using the chain rule). We do provide theoretical (next) and empirical arguments(Section~\ref{sec:object} and Section~\ref{sec:sentiment}) for its superiority to other methods from literature.

\subsection{Saturation of Neural Networks}
Saturation of neural networks is a central issue that hinders the understanding of nonlinear networks. This issue was discussed in details in~\cite{STY17} for Integrated Gradients. Basically, for a network, or a sub-network, even when the output crucially depends on some input, the gradient of the output w.r.t.~the input can be near-zero.

As an artificial example, suppose the network first transforms the input $x$ linearly to $y=2x$, and then transforms it to $z=\max(y, 1)$. Suppose the input is $x=1$ (where $z$ is saturated at value $1$), with $0$ being the baseline. Then for the hidden unit of $y$, gradient of $z$ w.r.t.~$y$ is $0$. Gradient*activation would be $0$ for $y$, which does not reflect the intuitive importance of $y$.
Like in Integrated Gradients, in computing conductance, we consider all extrapolated inputs for $x$ between $0$ and $1$, and look at the gradients of output w.r.t.~$y$ at these points. This takes the non-saturated region into account, and ends up attributing $1$ to $y$, as desired.

\subsection{Other methods can have wrong Polarity/Sensitivity}
\label{sec:polarity}

  In this section, we show that the methods we compare against can yield scores that have signs and magnitudes that are intuitively incorrect. This is intuitively because each misses terms/paths that our method considers.


  Activation values for a ReLU based network are always positive. However, ReLU nodes can have positive or negative influence on the output depending on the upstream weights.
  Here, Activation does not distinguish the sign of the influence, whereas condutance can.

  Gradient*Activation as a linear projection can overshoot.
  Certain hidden units that actually have near zero influence can be assigned high importance scores.
  For example, suppose that the network is the composition of two functions $f(x) = x$ and a weighted ReLU $g(y) = max(y-1,0)$.
  Again, the network computes the composition $g(f(x))$.   Suppose that the baseline is $x=0$ and the input is $x=1-\epsilon$. The output of the network is $0$.
  But the feature importance of the unit $f$ is deemed to be $1-\epsilon$ (activation) times $1$ (gradient), which is $1-\epsilon$.
  Notice that this is the only unit in its layer, so the fact that its influence does not agree in magnitude with the output is undesirable.
  In contrast, conductance assigns all hidden units a score of zero. The example can be extended to show that the feature importance score can disagree in sign with the actual direction of influence.

  Suppose that the network is the composition two functions $f(x) = -x$ and $g(y) = y$, i.e., the network computes the composition $g(f(x))$.
  Suppose that the baseline is $x=0$ and the input is $x=1$. The output of the network is $-1$. But the internal influence of the unit represented by the function $g$ is $+1$ (regardless of the choice of the input or the path).
  Notice that this is the only unit in its layer, so the fact that its influence does not agree in sign with the output is highly undesirable.
  In contrast, conductance assigns an influence score of $-1$.

  While these (bad) examples are caricatures, we find that these issues (of incorrect signs) do occur in real networks, for instance the sentiment analysis network from Section~\ref{sec:sentiment}).

\section{Applying Conductance to an Object Recognition Model}
\label{sec:object}

In this section, we describe our experiments in applying conductance to an
object recognition network. Such experiments also serve as empirical evaluations of
the effectiveness of conductance. The network is
built using the GoogleNet architecture~\cite{SLJSRAEVR14} and trained over the
ImageNet object recognition dataset~\cite{ILSVRC15}.
For a detailed description of the architecture, we refer the reader
to~\cite{SLJSRAEVR14}. We consider the following hidden layers in the network:
{\small \tt mixed3a, mixed3b, mixed4a, mixed4b, mixed4c, mixed4d,
mixed4e, mixed5a} and {\small \tt mixed5b}.

We consider the filters as a unit of analysis for the purpose of studying conductance.
The network we analyze was trained using ImageNet dataset. This dataset has
$1000$ labels with about $1000$ training images, and $50$ validation images per label.

We use conductance as a measure to identify influential filters in hidden
layers in the Inception network. Given an input image, we identify the top predicted label. For the
\emph{pre-softmax} score for this label, we compute the conductance for
each of the filters in each of the hidden layers using
Equation~\ref{eq:total-conductance}. For the visualization of the
conductance at pixel level, we use Equation~\ref{eq:pixel-conductance}. The
visualization is done by aggregating the conductance along the color channel
and scaling the pixels in the actual image by the conductance values.
See Figure~\ref{fig:selected_filters} for examples of images and filters that have high conductances for the image.

\begin{figure}[!htb]
    \tiny
    \begin{subfigure}{0.5\columnwidth}
  \includegraphics[width=0.5\columnwidth]{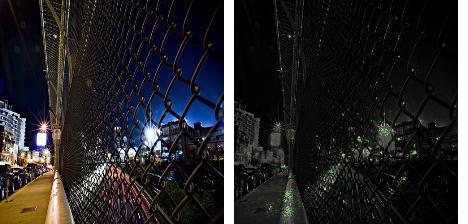}
    Chainlink fence \\

  \includegraphics[width=0.5\columnwidth]{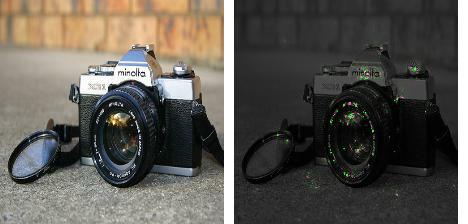}
    Camera \\
      \caption{{\small Filter $52$ in {\tt mixed3a} layer highlights glare.}}
    \end{subfigure}
    \begin{subfigure}{0.5\columnwidth}
  \includegraphics[width=0.5\columnwidth]{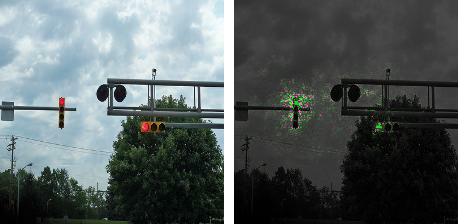}
  \begin{tabular}{ l }
    Traffic light \\
  \end{tabular}

  \includegraphics[width=0.5\columnwidth]{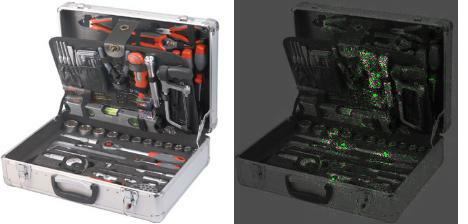}
  \begin{tabular}{ l }
    Carpenter's kit \\
  \end{tabular}
      \caption{{\small Filter $76$ in {\tt mixed4c} layer highlights red colored regions.}}
    \end{subfigure}
    \caption{{\small Some filters have high conductances across several images. Filter $52$ in {\tt mixed3a} layer highlights glare in first two photos,
    while filter $76$ in {\tt mixed4c} layer highlights red colored regions the
    3rd and 4th example.}}
   \label{fig:collisions}
  \end{figure}

Despite the fact that relatively few filters determine the prediction of the
network for each input, we found that
some of the filters are shared by more than one images with different labels.
Two examples of such filters are shown in Figure~\ref{fig:collisions}.

\subsection{Ablation Study}
\label{sec:object-eval}
We sanity check the conductance of filters by comparing them with the drop in
the pre-softmax score for the top predicted label if we ablate the filters. We
  call this the ablation score for a filter. \footnote{Ablating a filter is
    done by setting the bias term to $-\infty$.}. Because we are going to ablate the filters
  one by one, this is pretty much like computing a discrete gradient. So it is going to be artifactually similar to gradient*ablation.
  However, this eval will separate internal-influence and activations from conductance and gradient*activation.


  We used a sample of $100$ images from ImageNet validation data. For
  each image, we computed top $10$ filters based on conductance. For each of
  these filters we computed its ablation score. (Computing conductance in
  tensorflow involved adding several gradient operators and didn't
  scale very well.) Hence we limited our analysis to only $100$ images. We repeated this process using internal influence, activations and gradient*activation.
  We report the Pearson correlation between each of these measures and the ablation scores.
  Conductance has a correlation of $0.79$, internal influence has a correlation of $0.487$ (25th percentile = $0.38$,
75th percentile = $0.61$), activations had a correlation of $0.07$, and gradient*activations had a correlation score of $0.92$.

Next we studied how many filters we need to ablate in the network in order
for the network to change its prediction. We found that, it is sufficient to
ablate $3.7$ on an average for the network to change its prediction for an
image. Only $3$ out of $100$ images needed more than $10$ filter ablations to
change the predicted label. The maximum was $16$. This provides further
evidence that using conductance we can identify filters that are important for
the prediction.

We compare this to the filters with highest internal influence. Out of the
$100$ sample images, the network prediction changed for only $5$ images
when their top 10 filters (according to internal influence) were ablated.





\subsection{Feature Selection Study}
Our second empirical evaluation (ablations being the first) for feature importance to
by selecting features that are important for a class/label. A good feature
importance method should be able to identify the features important to not just
an input instance, but to an input class. That is to make statements of the
form: “yellowness is a salient feature of all bananas”, or “wheels are features
of all cars”. This eval is somewhat similar to the one in~\cite{QVL13}, which also builds classifiers from high-level
concepts. They do it using an autoencoder with sparsity. Whereas our
formulation treats filters in hidden layers as representing high-level concepts.

In our study, we chose four sets of five labels each from ImageNet. The first two
sets were thematically similar (5 species of dog and 5 types of water vessels)
and the other two sets had labels chosen randomly. We picked about 50 images per label from the
validation set -- we split this into 30 images for training and 20 for eval. For each method of feature importance, we compute an importance value for each
filter using the method and aggregated those over training set per label.
We pick $k$ filters with highest aggregate value for any label.
We then use these $k$ filters to classify images from eval set, by training
a simple linear classifier. A method that produces the right notion of feature
importance should result in a better predictor because it identifies a better
list of features. We have displayed the results for four methods for these four
sets of labels in Table~\ref{table:feature_selection}. We report results for
two label sets: 5 types of water vessels as well as one random label set, as
well as aggregate numbers over the four different sets of labels.
We observe that conductance leads to better feature selection from these results.

  \begin{table}
    \tiny

\begin{tabular}{lllll}
  \hline
  Method & $5$ features & $10$ features & $15$ features & $20$ features \\

\hline

activations & 37.80 & 36.59 & 47.56 & 54.88  \\
gradient*activation & 32.93 & 58.54 & 64.63 & 69.51 \\
influence & 45.12 & 53.66 & \textbf{71.95} & 68.29  \\
conductance & \textbf{48.78} & \textbf{68.29} & 70.73 & \textbf{79.27}  \\
\hline

  \end{tabular}
Water vessels task

  \begin{tabular}{lllll}
  \hline
  Method & $5$ features & $10$ features & $15$ features & $20$ features \\
\hline
activations & 54.76 & 61.90 & 65.48 & 69.05 \\
gradient*activation & 67.86 & 83.33 & 85.71 & 88.10 \\
influence & 50.00 & 67.86 & 73.81 & 75.00 \\
conductance & \textbf{88.10} & \textbf{94.05} & \textbf{96.43} & \textbf{94.05} \\

\hline

  \end{tabular}
Random labels task.
\quad

  \begin{tabular}{lllll}
  \hline
  Method & $5$ features & $10$ features & $15$ features & $20$ features \\

\hline

activations & 46.52 & 48.51 & 55.81 & 60.45 \\
gradient*activation & 56.94 & 76.01 & 79.74 & 84.39 \\
influence & 48.50 & 62.23 & 71.37 & 72.50 \\
conductance & \textbf{68.85} & \textbf{81.58} & \textbf{85.99} & \textbf{87.79} \\

\hline
  \end{tabular}
Aggregate over 4 tasks.

    \caption{{\small Accuracy of classifiers trained on small number of features
    selected using the four different methods.
    The first table reports accuracy for classifying between water vessals. The
    second table shows numbers for classifying between 5 randomly chosen
    labels (matchstick, meerkat, ruffed grouse, cheeseburger, toaster).
    The last table reports aggregate over all four label sets.
    }}
\label{table:feature_selection}
\end{table}

\begin{figure}[!htb]
    \tiny
    \begin{subfigure}{0.5\columnwidth}
  \includegraphics[width=0.5\columnwidth]{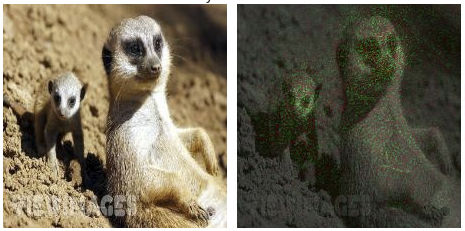}
  meerkat \\
  \includegraphics[width=0.5\columnwidth]{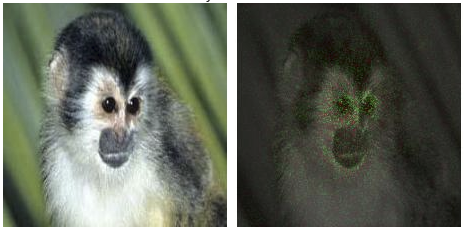}
  titi \\
  \caption{{\small Filter $757$ in {\tt mixed5a} layer selected for meerkat label also
  has high conductance for the titi.}}
    \end{subfigure}
    \begin{subfigure}{0.5\columnwidth}
  \includegraphics[width=0.5\columnwidth]{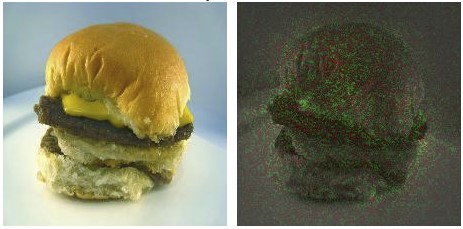}
  cheeseburger \\
  \includegraphics[width=0.5\columnwidth]{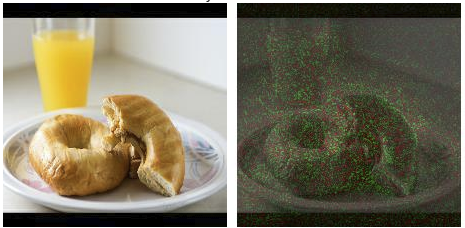}
  bagel \\
    \caption{{\small Filter $7517$ in {\tt mixed5a} layer selected for cheeseburger label also
  has high conductance for the bagel.}}
    \end{subfigure}
    \caption{{\small Example images from imagenet that have high conductance for
        two filters that were important for the labels meerkat and cheeseburger.
        These images were selected from outside the set of images that were used to identify these filters as important to these classes.}}
   \label{fig:selected_filters}
  \end{figure}

  In Figure~\ref{fig:selected_filters}, we present exemplar images \emph{outside} the training set
  that have high conductance for two filters that were identified as important for the labels meerkat and cheeseburger respectively.
  As a sanity check, Titis and Bagels are indeed similar to meerkats and cheeseburgers respectively.

\section{Applying Conductance to a Sentiment Model}
\label{sec:sentiment}

  Next we analyze a model that scores paragraphs or sentences for sentiment. The model is a convolutional model from~\cite{sentiment} trained over review data.
  In the variant of the model we study, the words are embedded into a $50$ dimensional space. 
  The embeddings are fed to a convolutional layer that has $4$ window sizes ranging from $3$ to $6$ words. Each of the four filter widths has $64$ feature maps.
  Each of the $4\times64 = 256$ filters is $1$-max pooled to construct a layer of size $256$, which is then fed (fully connected) to a layer of $10$ neurons, which is fed to a logit, that feeds into a softmax that classifies whether the sentiment is positive and/or negative (two binary heads).

  We study the conductances of the filters in the convolutional layer. We study
  the how filters collaborate in section~\ref{sec:division}. In
  section~\ref{sec:negation}, we look at how positive and negative sentiments
  are captured.


  
\subsection{Ablation Study}
\label{sec:sentiment-eval}
  We now compare the conductance of a filter to the change in score due to ablating the filter.
  We do this over a sample of $100$ inputs drawn at random from the Stanford Sentiment Tree Bank~\cite{stanford-tree}.
  We produce a similar comparison between internal influence.

  We observe that there is good agreement between conductance and the ablation scores of the filters.
  The average (over the inputs) Pearson correlation is $0.88$.
  The correlation between activations and the ablation scores is $0.19$ and between $gradient*activations$ and the ablation scores it is 0.99 (as we discuss in Section~\ref{sec:object-eval}, the high correlation is somewhat artifactual.).

  In contrast, the internal influences and the ablation scores are negatively correlated.
  The average (over the inputs) Pearson correlation is $-0.43$. This is not surprising as the internal influence definition misses two scaling terms from the conductance formula, which can alter signs. In contrast, recall from Section~\ref{sec:properties} that conductances satisfy completeness, therefore they are bound to have the right scale.
  The negative correlation is explained by the fact that one of the two scaling missing terms (specifically the embedding vector) has predominantly negative values.
  Notice that internal influence had a relatively better Pearson correlation for the object recognition model (compare to Section~\ref{sec:object-eval}). The effect of the lack of scaling terms varies across networks. It does not appear that internal influence~\cite{influence} was applied to a text task, or perhaps they did not notice the issue. Our analysis highlights the importance of the scaling terms and the completeness property.

  Another popular way of understanding the importance of hidden neurons is to study their activation.
  This is misleading for similar reasons. For a ReLU based network, activations are always positive. Whereas, neurons may either have positive or negative influence. This can happen even within a single input.
  In supplementary material (Figure 1) we show a plot of the ratio of the
  absolute value of the sum (over filters) of ablation scores to the sum of the
  absolute values of the ablation scores for a $100$ inputs. If the ablation
  scores across filters for a given input all agreed in sign, this ratio would
  be $1.0$. Notice that most of the ratios are far less than $1.0$, indicating
  that the final prediction score is the sum total of several positive and
  negative effects.

\subsection{Division of labour}
\label{sec:division}

\begin{figure}[!htb]
    \centering
   \includegraphics[width=0.4\columnwidth,height=0.2\columnwidth]{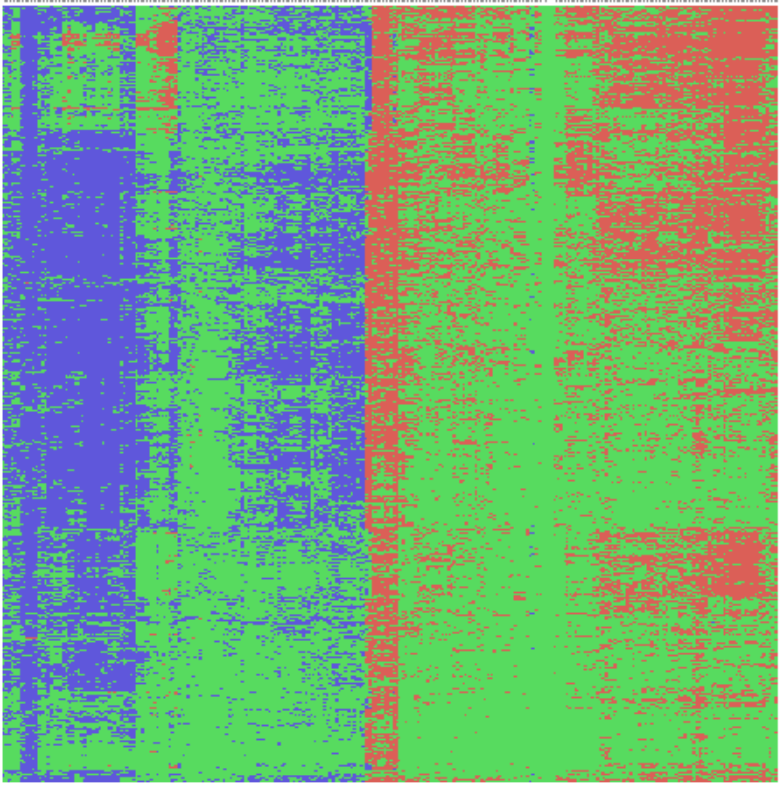}
    \caption{{\small Visualization of signs of conductances for around 4000 examples (rows) by the $256$ fitlers (columns). We color negative conductance with red, positive conductance with blue and conductance near zero (we apply a threshold of $0.01$ for numerical stability) with green. Notice that most filters exclusively capture either negative sentiment or positive sentiment.}}
    \label{fig:heat}
  \end{figure}

We notice that almost all the filters either capture positive sentiment or negative sentiment, but not both.
We substantiate via Figure~\ref{fig:heat}, which is a  clustered heatmap of signs of conductances of the $256$ filters (columns) for around four thousand examples (rows) from the Stanford Sentiment Tree Bank~\cite{stanford-tree}.
Notice that very few filters have both negative and positive conductance.

\subsection{Negation}
\label{sec:negation}

\begin{figure}[!htb]
    \centering
   \includegraphics[width=0.6\columnwidth]{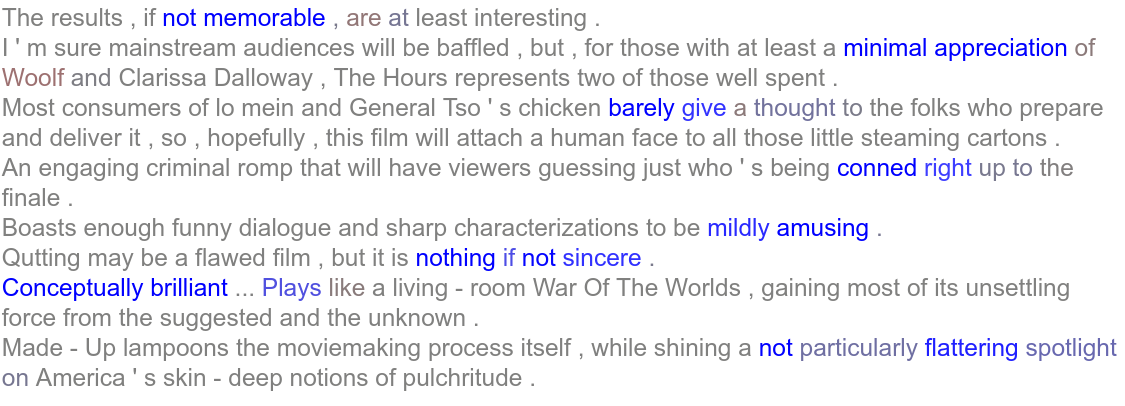}
    \caption{{\small Sentences with high conductance for filters that have high conductance for the phrase ``not good''. These filters capture negation, diminishing, and show some stray errors.}}
    \label{fig:not-good}
  \end{figure}

\begin{figure}[!htb]
    \centering
   \includegraphics[width=0.6\columnwidth]{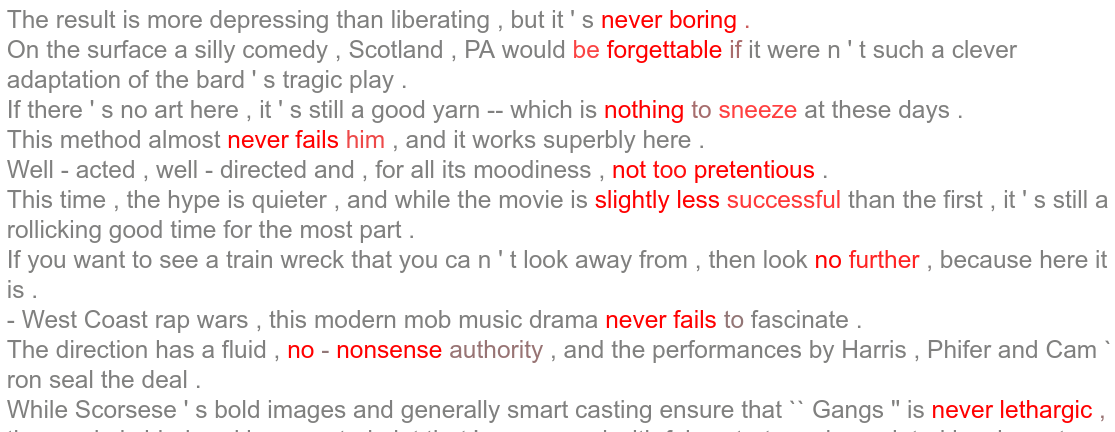}
      \caption{{\small Sentences with high conductance for filters that have high conductance for the phrase ``not bad''. These filters are largerly focussed on negation. }}
    \label{fig:not-bad}
  \end{figure}

Negation is commonly used in expressing sentiments, in phrases like ``this is not good'' or ``this is not bad''.
Does the sentiment network understand negation? Does it have hidden units dedicated to implement the logic of negation?
We first identify high conductance filters for the input ``this is not good'' that have a high attribution to the pattern ``not good''.
We find that three filters fit this pattern. It is possible that these filters perform different functions over other inputs.
To show that this is not the case, we then run the network over around $4000$ inputs from the Stanford Sentiment Treebank, noting examples that have a high conductance for any of these three filters.
Figure~\ref{fig:not-good} displays the results. Figure~\ref{fig:not-bad} has analogous results for the pattern ``not bad''.
In the second case, the filters seem quite focused on negation. This suggests that the network does understand negations,
and has dedicated filters for handling negations.

\bibliography{ms}
\bibliographystyle{unsrt}

\end{document}



\maketitle

\section*{Appendix}

\begin{figure}[!htb]
  \centering
  \includegraphics[width=0.46\columnwidth]{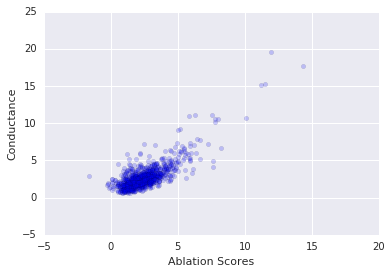}
  \includegraphics[width=0.48\columnwidth]{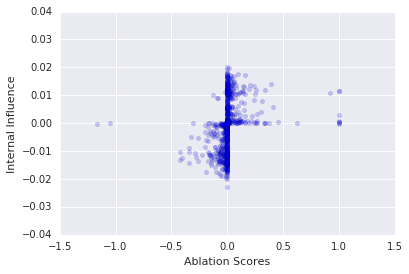}
  \caption{{\small Left panel: conductance vs. ablation scores. Right Panel: internal influence vs. ablation scores.
   There is one data point per image, filter combination. Ablations scores show a stronger linear relation with the conductances.}}
\label{fig:scatter-conductance}
\end{figure}
